\documentclass[accepted]{uai2021} 

\usepackage[american]{babel}

\usepackage{natbib} 
\usepackage{booktabs} 
\usepackage{tikz} 

\DeclareMathAlphabet{\mathcal}{OMS}{cmsy}{m}{n}

\usepackage[utf8]{inputenc} 
\usepackage[T1]{fontenc}    
\usepackage{url}            
\usepackage{booktabs}       
\usepackage{amsfonts}       
\usepackage{nicefrac}       
\usepackage{microtype}      

\usepackage{bbm}
\usepackage{amsmath, amssymb}
\usepackage{enumitem}
\usepackage{xcolor}

\usepackage{graphicx}
\usepackage{amsthm}
\usepackage{multirow}
\usepackage{todonotes}
\usepackage{bm}

\newtheorem{theorem}{Theorem}
\newtheorem{lemma}{Lemma}

\usepackage{mwe}
\usepackage{array}
\usepackage{todonotes}
\usepackage{textcomp}

\DeclareMathOperator*{\argmax}{arg\,max}

\newlength{\tempdima}

\newcommand{\rowname}[1]
{\rotatebox{90}{\makebox[\tempdima][c]{\textbf{#1}}}}

\hypersetup{colorlinks,citecolor=blue}

\title{Statistically Robust Neural Network Classification}

\author[1]{\href{mailto:Benjie Wang <benjie.wang@cs.ox.ac.uk>?Subject=SRNNC Paper}{\color{black} Benjie~Wang}{}\thanks{Contact Author: benjie.wang@cs.ox.ac.uk}}
\author[2]{Stefan~Webb}
\author[3]{Tom~Rainforth}
\affil[1]{%
    Department of Computer Science\\
    University of Oxford
}
\affil[2]{%
    Twitter Cortex\\
    San Francisco
}
\affil[3]{%
    Department of Statistics\\
    University of Oxford
}

\begin{document}
\maketitle

\begin{abstract}
Despite their numerous successes, there are many scenarios where adversarial risk metrics do not provide an appropriate measure of robustness. For example, test-time perturbations may occur in a probabilistic manner rather than being generated by an explicit adversary, while the poor train--test generalization of adversarial metrics can limit their usage to simple problems.
Motivated by this, we develop a probabilistic robust risk framework, the \textit{\textbf{statistically robust risk}}
 (SRR), which considers pointwise corruption distributions, as opposed to worst-case adversaries. The SRR provides a distinct and complementary measure of robust performance, compared to natural and adversarial risk. We show that the SRR admits estimation and training schemes which are as simple and efficient as for the natural risk: these simply require noising the inputs, but with a principled derivation for exactly how and why this should be done.
 Furthermore, we demonstrate both theoretically and experimentally that it can provide superior generalization performance compared with adversarial risks, enabling application to high-dimensional datasets.
\end{abstract}

\section{INTRODUCTION}

Since the discovery of the phenomenon of adversarial examples for neural networks \citep{SzegedyEtAl2014,GoodfellowEtAl2015,papernot2016limitations}, a variety of approaches for assessing and mitigating their impact on decision-making systems have been proposed~\citep{gu2014towards,moosavi2016deepfool,MadryEtAl2018}. Much of this work has focused on the formal verification of neural network classifiers, such as the robustness of predictions under a $L_p$-norm perturbation set \citep{GehrEtAl2018, WangEtAl2018}, typically doing this in an input specific manner.
Motivated by explicit adversarial attacks, these approaches are focused on worst-case robustness: they are based on the largest loss within the perturbed region.

Though highly appropriate in a variety of cases, this general approach is not universally applicable.
Firstly,  one is often concerned about robustness to naturally occurring, or \textbf{\emph{random}}, input perturbations, rather than an explicit adversary.
For example, in self-driving cars we may not have access to the exact inputs due to sensor imperfections and wish to ensure our predictions are robust to such variations.
Here our classifier must account for these variations, but some level of risk will usually be acceptable: it will typically be neither feasible nor necessary to guarantee there are \emph{\textbf{no}} possible adversarial inputs, but we instead wish to ensure the \emph{\textbf{probability}} of encountering such an input is sufficiently low.

Secondly, in practice, one is usually concerned with the \emph{\textbf{overall}} robustness of the network, that is, its robustness across the range of possible inputs that it will see at test-time.
This has motivated network-wide worst-case robustness definitions, such as average minimal adversarial distance \citep{FawziEtAl2018} and adversarial risk \citep{MadryEtAl2018}, along with associated training schemes~\citep{WongKolter2018,MadryEtAl2018}.
However, whereas the motivation for requiring worst-case robustness for individual inputs is often clear, it is more difficult to motivate using worst-case robustness for the classifier as a whole;
a classifier can only be perfectly worst-case robust if it is robust to all possible perturbations of all inputs, something which will very rarely be achievable in practice.
Moreover, previous work has shown that worst-case robustness metrics can have very poor generalization from train to test time, both theoretically and in practice for real networks, substantially reducing their applicability~\citep{SchmidtEtAl2018,YinEtAl2019}.

To address these limitations, we suggest a class of alternative robust risk metrics, which we term \textbf{\textit{statistically robust risks}} (SRRs), that naturally arise when relaxing worst-case adversaries to pointwise perturbation distributions. 
SRRs can be used to assess the overall probabilistic robustness of a classifier by averaging a loss function over both possible inputs and an input perturbation distribution.
Unlike adversarial risks, our SRR framework naturally applies at a network-wide level due to the law of total expectation.
We emphasize that this framework is not a replacement for adversarial risk, or a means to learn adversarially robust networks, but a distinct and complementary measure of robustness that will be more appropriate in some scenarios.
Our work can be viewed as an extension and generalization of the pointwise statistical robustness work of~\citet{WebbEtAl2019}, which quantifies the expected robustness of an \emph{\textbf{individual}} datapoint under a perturbation distribution.

\paragraph{Contributions}

Our contributions are as follows. Firstly, we provide theoretical and empirical results showing that SRRs have superior generalization performance to their corresponding adversarial risks, particularly in high-dimensions, with bounds on the generalization error respectively scaling as $O(\log(d))$ and $O(\sqrt{d})$ in the size of the network. This suggests that it may be possible to obtain statistically robust networks in a wide range of applications where adversarial robustness is still elusive or inappropriate. Further, we show that the SRR admits efficient estimation and training schemes which incur no extra computational cost over standard training. Indeed, training to a SRR requires only a noising of the inputs passed to the network, such that it encompasses, motivates, and formalizes many commonly-used heuristics.

We justify the practical utility of our statistical robustness metric with a number of novel insights. Firstly, we demonstrate that SRRs can differ significantly from both their corresponding natural (i.e.,~non-robust) and adversarial risk, and as such provide a unique metric for both training and testing networks that helps ensure robustness to probabilistic input perturbations. Secondly, we find that SRRs generalize well \textit{across different perturbation distributions}, meaning that it is not necessary to have knowledge of the precise test-time perturbation distribution. Finally, we show that practical safety properties encoded through bespoke loss functions can be tackled through SRRs, while standard training suffers from overfitting and instability.

\section{BACKGROUND}

\subsection{Adversarial Examples}
Although the general concept of adversarial examples (a perturbed input data point that is classified poorly) is well understood, the precise definition is often left implicit in the literature, despite many versions being present~\citep{DiochnosEtAl2018}. To formalize this, let $f_\theta$ represent the classifier (with parameters $\theta$) and $c$ label the true class. Let $x$ be the original input point and $x'$ the perturbed input point. 
Three different definitions of an adversarial example are now commonly used:
\begin{itemize}
	\item \textbf{Prediction change (PC)} $f_\theta(x') \neq f_\theta(x)$;
	\item \textbf{Corrupted instance (CI)} $f_\theta(x') \neq c(x)$;
	\item \textbf{Error region (ER)} $f_\theta(x') \neq c(x')$.
\end{itemize}

The distinction between PC and CI is that the former is concerned with whether a perturbation changes the classification (regardless of whether it is correct), while the latter concerns whether the perturbed point is classified correctly. 
The ER definition is typically not measurable, since we usually will not have labels for perturbed points $x'$.
Since we are interested in risk metrics, we take the CI definition.

\subsection{Natural and Adversarial Risks}
\label{sec:adv_risk}
The \textit{\textbf{risk}} of a classifier $f_\theta$ is a measure of its average performance with respect to the data distribution:
\begin{align}
\label{eq:risk}
r_\mathcal{D}(f_\theta) &\triangleq \mathbb{E}_{(X, Y) \sim p_\mathcal{D}} \left[L(X, Y, f_\theta) \right],
\end{align}
where $(X, Y)$ is an input/target pair, $p_\mathcal{D}$ is the true data generating distribution, and $L$ is some loss function. For non-robust risks, $L(X, Y, f_{\theta})$ can usually be written in the form $\phi(f_\theta(X), Y)$, which we term the \emph{\textbf{natural risk}}. 
An empirical version of this, $R_N(f_\theta)$, can be obtained by replacing the expectation with a Monte Carlo average over a training dataset $\{(x_1, y_1), ..., (x_N, y_N)\}$. 
Training a classifier then corresponds to solving the optimization problem $\min_\theta R_N(f_\theta)$.

To model the effect of an adversary limited to additively perturbing inputs by a vector $\delta$ within a limited set $\Delta$ (e.g., an $L_\infty$-ball), \textit{\textbf{adversarial risk}} is defined as
\begin{align}
r^{\text{adv}}_\mathcal{D}(f_\theta) &\triangleq \mathbb{E}_{(X, Y) \sim p_\mathcal{D}} \left[\max_{\delta \in \Delta}\phi(f_\theta(X+\delta), Y) \right],
\end{align}
which is in fact a form of risk with loss function $L^{\text{adv}} \triangleq \max_{\delta \in \Delta}\phi(f_\theta(X+\delta), Y)$. When the 0-1 loss function $\phi(f_{\theta}(X),Y) = \mathbbm{1}_{\arg \max_{i = 1,\cdots, M} f_\theta(X)_i \neq Y}$ is used, this is known as \emph{\textbf{adversarial accuracy}}.
Optimizing the empirical adversarial risk, $R_N^{\text{adv}}(f_\theta)$,
corresponds to a {{robust optimization}} problem \citep{BenTalEtAl2009}. 
\textit{\textbf{Adversarial training}} \citep{GoodfellowEtAl2015, KurakinEtAl2018, MadryEtAl2018} solves the problem by lower bounding the inner maximization using gradient-based methods to generate maximally adversarial examples, and training on this approximate loss.

\subsection{Statistical Robustness} \label{statrob}

\citet{WebbEtAl2019} recently introduced a \textit{{statistical robustness}} metric that provides a probabilistic alternative to formal verification of \emph{\textbf{pointwise}} robustness. Standard verification schemes target the binary 0-1 metric on whether an adversarial example exists in a {{perturbation region}} $\Delta$ around a point. Their statistical robustness metric instead corresponds to the \textbf{\emph{probability}} of drawing an adversarial example from some \emph{\textbf{perturbation distribution}}. Concretely, for a perturbation distribution $p(\cdot|x)$ centred around $x$, they define their statistical robustness metric as
\begin{align} \label{eq:psr}
\begin{split}
\mathcal{I}[p] &\triangleq 
\mathbb{E}_{X' \sim p(\cdot|x)} \left[\mathbbm{1}_{f_\theta(X') \neq f_\theta(x)}\right].
\end{split}
\end{align}
This generalizes, and provides more information than, verification about the network's robustness around $x$: if there is no adversarial example in the support of $p(\cdot|x)$, then the probability is 0, whereas if there is an adversarial example, the metric indicates how likely we are to encounter one.

\subsection{Generalization}
The problem of generalization is fundamental in machine learning: we want classifiers to perform well on not just training points but unseen test points. It is well known in statistical learning theory that we can probabilistically upper bound the generalization error $r_\mathcal{D}(f) - R_N(f)$ of a learning algorithm using notions of complexity on the admissible set of classifiers (e.g. all parameterizations of a neural network) and loss function \citep{ShalevShwartzBenDavid2014}. Intuitively, if the admissible set of functions is less complex, then there is less capacity to overfit to the training data.

To be more precise, we define the \textit{\textbf{empirical Rademacher complexity}} (ERC) for function class $\mathcal{F}: \mathbb{R}^d \to \mathbb{R}$ and sample set $\mathcal{S} = \{x_1, ..., x_N\}, \, x_i\in\mathbb{R}^d$ to be \citep{ShalevShwartzBenDavid2014}:
\begin{align}
\label{eq:emp_rad}
	\textup{Rad}_\mathcal{S}(\mathcal{F}) := \frac{1}{N} \mathbb{E}_\sigma \left[\sup_{f \in \mathcal{F}} \sum_{n=1}^{N} \sigma_n f(x_n)\right],
\end{align}
where $\sigma_1, ..., \sigma_N$ are independent Rademacher random variables, which take either the value $-1$ or $+1$, each with probability $1/2$. Intuitively, this measures the complexity of the class by determining how many different ways functions $f \in \mathcal{F}$ can classify the sample $\mathcal{S}$.

\section{FROM ADVERSARIAL TO STATISTICAL RISKS}

Adversarial examples originally captured the attention of the machine learning community by demonstrating a discrepancy between the behaviour of NNs and human reasoning. Given that, in domains such as computer vision and natural language processing, the long-term goal is to attain models which can reason as humans do, it is natural to define robustness in terms of all semantically meaningful perturbations. However, given the rapid adoption of machine learning systems in applications such as autonomous vehicles and medical diagnosis, robustness is now also a vital practical requirement: brittleness to input perturbations can have severe consequences. These goals, 
while seemingly aligned, can in fact sometimes be conflicting. We argue that the latter agenda requires its own treatment and develop an associated robustness framework that arises naturally when considering how ideas from adversarial robustness can be transferred to probabilistic settings.

\subsection{The Shortfalls of Adversarial Approaches} \label{sec:shortfall}

Obtaining \textbf{\textit{fully}} adversarially robust networks (in the sense of being robust to all meaningful perturbations to all possible points) is a typically infeasible task, even for simple datasets such as MNIST and when the set of semantically consistent perturbations is known~\citep{SchottEtAl2019,SchmidtEtAl2018,YinEtAl2019}.
As such, one must rely on robustness metrics, such as adversarial risk/accuracy. 

However, this can have significant issues. First, adversarial risk loses information about \textbf{\textit{how}} robust a point is. By taking the worst point within a perturbation set, adversarial risk is by definition independent of performance on the vast majority of the perturbed input space (so long as it is better than the worst point). Adversarial risk thus places stringent $0/1$ requirements on each point from the data distribution, such that it favors a greater \textit{number} of adversarially (completely) robust points, without guarantees of any degree of robustness on other points. When considering applications where perturbations are randomly generated, this can be very misleading, or even dangerous. 

Consider, for instance, a network trained to classify disease based on medical imaging. Due to imperfections in the imaging equipment, as well as variation in equipment across different hospitals, random noise may be introduced to the test dataset. Optimising for adversarial risk might make the network adversarially robust on 80\% of test points; however, if it is often fooled by random noise on the remaining 20\%, this could result in misdiagnoses for many patients. In contrast, a network is robust against 95\% of random perturbations overall would be preferable, even if it is adversarially robust on many fewer points. In these cases, somewhat counterintuitively, adversarial risk does not correspond well with the required notion of safety or robustness.

Second, learning with adversarial risk has proven to be very difficult from a statistical learning perspective due to poor generalization. Though adversarial training is effective for reducing the adversarial risk of neural networks on small datasets like MNIST, success has been limited in scaling up to higher-dimensional datasets. \citet{SchmidtEtAl2018} show this is due to a generalization gap, whereby, for CIFAR-10, it is possible to achieve adversarial accuracy of 97\% on the training set, yet just 47\% on the test set. This overfitting is in contrast to the natural case, where well-tuned networks rarely overfit on CIFAR-10~\citep{YinEtAl2019}.

In view of this, in the practical robustness agenda there is thus a need for robustness metrics which are not as conservative as adversarial risk, taking into account performance on a larger subset of input space, whilst also being amenable to training and tractable in the sense of generalization.

\subsection{Total Statistical Robustness Metric} \label{sec:tsrm}

The pointwise statistical robustness framework introduced in Section~\ref{statrob} provides an indication of how we might construct some form of robust statistical risk.
However, it is not directly applicable as a) we require a mechanism for assessing the \emph{\textbf{overall}} robustness of a network; b) it only examines \emph{\textbf{changes}} in predictions (PC), such that it can be satisfied by trivial networks which always make the same incorrect prediction; and c) it does not provide any practical mechanism for \emph{\textbf{training}} networks.

To address the first two issues, we now introduce the \textit{\textbf{total statistical robustness metric}} (TSRM) as follows:
\begin{align}
\label{eq:total}
\mathcal{I}_{\text{total}}[p]
&= \mathbb{E}_{(X, Y) \sim p_\mathcal{D}}\left[\mathbb{E}_{X' \sim p(\cdot|X)}\left[\mathbbm{1}_{f_\theta(X') \neq Y}\right]\right],
\end{align}  
where $p_\mathcal{D}$ is the true data generating distribution and $Y = c(X)$ is the true label of $X$. 
As it includes an expectation over the data, the TSRM is a measure for the overall robustness of the network.
Intuitively, it can be thought of as the classification error under test-time input corruptions $p(\cdot|X)$.

Notice that the TSRM also varies from~\eqref{eq:psr} in that it compares $f_\theta(X')$ to $Y$ instead of $f_\theta(X)$. That is, we now consider CI examples.
This is because we want to assess how the network performs over distribution $\mathcal{D}$. Analogously, while pointwise adversarial metrics are usually defined in terms of prediction change (PC), adversarial accuracy is defined in terms of misclassification (CI).

\subsection{Statistically Robust Risk} \label{srrk}

The TSRM forms a useful metric for pre-trained networks, but it is not suitable as a training objective due to the difficulty of taking gradients through the identity function.  
Further, it explicitly assumes we are interested in probability of failure, rather than a more general loss.
To address this, we note that it can be thought of as a specific risk and thus generalized to
\begin{align} \label{rstat}
\hspace{-6pt}r^{\text{stat}}_\mathcal{D}(f_\theta) \triangleq \mathbb{E}_{(X, Y) \sim p_\mathcal{D}}\left[\mathbb{E}_{X' \sim p(\cdot|X)}\left[\phi(f_\theta(X'), Y) \right]\right]
\end{align}
where $\phi$ represents a natural, pointwise, loss function as per Section~\ref{sec:adv_risk}.\footnote{In certain cases, we may further require $\phi$ to also take $X$ directly as an input. This potential dependency is not problematic and omitted simply for notational clarity.}
We refer to $r^{\text{stat}}_\mathcal{D}(f_\theta)$ as a \textit{\textbf{statistically robust risk}} (SRR).
The TSRM now constitutes the special case of $\phi(f_{\theta}(X'),Y)=\mathbbm{1}_{f_\theta(X') \neq Y}$.
Note that the SRR corresponds to using the loss function
\begin{align} \label{eq:lstat}
L^{\text{stat}}(X, Y, f_\theta) &\triangleq \mathbb{E}_{p(X'|X)}\left[\phi(f_\theta(X'), Y) \right].
\end{align}
Given a dataset $\{(x_1, y_1), (x_2, y_2), ..., (x_N, y_N)\}$, we can also define the \textit{\textbf{empirical SRR}}:
\begin{align}
\label{eq:exact_emp_rstat}
R^{\text{stat}}_{N}(f_\theta) \triangleq \frac{1}{N} \sum_{n = 1}^{N} \mathbb{E}_{p(X'|x_n)}\left[\phi(f_\theta(X'), y_n) \right].
\end{align}
The SRR framework provides a mechanism for linking statistical robustness back to the conventional notions of natural and adversarial risk, as well as a basis for theoretical analysis (see Section~\ref{genz}).
The natural risk can be viewed as a special case of a SRR for which $p(X'|X)$ collapses to a Dirac delta measure about $X$, such that it does not take into account robustness to perturbations.
On the other hand, by using the expected loss over $p(\cdot|X)$, instead of just the single worst-classified perturbation, a SRR contains important information that is not captured by an adversarial risk.
We illustrate in Appendix \ref{apx:effect} how this affects the behaviour of classifiers trained on SRR, in particular, improving the pointwise statistical robustness of a greater number of points.

At a high-level, training using a SRR has the effect of ``smoothing'' the decision boundaries relative to using the corresponding natural risk.
This can be useful when we want to be sensitive to certain classes or events, as it allows us to train our classifier to take conservative actions when the input is close to potentially problematic regions.
For example, a self-driving car needs to ensure it avoids false negatives when predicting the presence of a pedestrian.

\subsection{Estimation and Training} \label{esttra}

The empirical SRR cannot be evaluated exactly as it contains an expectation over a perturbation distribution. A simple approximation approach is to use Monte Carlo estimation for each inner expectation, that is:
\begin{align} \label{eq:MC}
	R^{\text{statMC}}_{N, C}(f_\theta) = \frac{1}{N} \sum_{n=1}^{N} \frac{1}{C} \sum_{m=1}^{C} \phi(f_\theta(x'_{n, m}), y_n) 
\end{align}
where $\{x'_{n, m}\}$ is a sample from the perturbation distribution $p(\cdot|x_n)$ for $m \in \{1,\dots, C\}$.
Estimating $\mathbb{E}_{p(X'|x_n)}\left[\phi(f_\theta(X'), y_n) \right]$ in this way can be a challenging task when this expectation is very small (i.e. when the point is very robust), 
typically requiring specialist rare-event estimation methodology to avoid large \textbf{\textit{relative errors}} \citep{WebbEtAl2019}. 

Perhaps surprisingly though, this difficulty actually resolves itself when considering the empirical SRR as an overall estimation problem.
This is firstly because, for practical networks and tasks, the empirical SRR is typically dominated by a small subset of the inputs $x_n$ for which the pointwise statistical robustness loss is large (``non-robust'' points), as opposed to the majority of inputs where the pointwise statistical robustness loss is very small (``robust'' points). 
Consequently, large relative errors for these robust points do not significantly impact on the overall error.
Secondly, we usually have relatively large datasets ($N > 10^5$) for tasks involving neural networks, meaning that we do not necessarily need accurate estimates of pointwise robustness around each individual datapoint to obtain an accurate average: by the law of large numbers, the errors in our (unbiased) estimates will cancel each other when averaged.

Because of these effects, we found that Monte Carlo estimation with $C = 1$ (sampling a single point $x'_n$ from $p(\cdot | x_n)$, then averaging the loss over these points) is often sufficiently accurate in practice for training and evaluation.	

To use the SRR as a \emph{\textbf{training objective}}, we need to use a differentiable loss $\phi$, such as the cross-entropy, so that we can perform stochastic gradient descent.
We can then iterate through the training data by taking mini-batches $B\subset \{1,\dots,N\}$, drawing corresponding sample perturbations $x_{n, m}'\sim p(\cdot | X=x_n)$, and updating the network using the unbiased gradient updates
\begin{align} \label{grad}
\nabla_{\theta} r^{\text{stat}}(p,f_\theta) \approx \frac{1}{\|B\|} \nabla_{\theta} \sum_{n\in B} \frac{1}{C} \sum_{m = 1}^{C} \phi(f_{\theta}(x_{n, m}'),y_n),
\end{align}
noting that this is equivalent to conventional training but with the inputs randomly perturbed.

These insights mean that we can estimate and train on SRRs accurately with no additional cost to standard neural network training. This also provides justification for data augmentation schemes which sample a single perturbation for each datapoint, as accurately minimizing a statistically robust risk (without needing to sample many different perturbations, or use the original datapoint). 
Though this scheme is simple and efficient, our framework is itself agnostic to how we estimate/train (see Appendix \ref{apx:esttra} for discussion).

\subsection{Choice of Perturbation Distribution} \label{cpd}
The aim of the perturbation distribution used in a SRR is not to be a completely accurate model of test-time perturbations; finding such a perturbation distribution is typically infeasible. 
Instead, it should be chosen to reflect what kind of perturbations we wish to be robust to. 
For example, if we are concerned about large-magnitude perturbations, we might use a heavy-tailed distribution such as a Cauchy.
We may instead have known invariances in our inputs (e.g.~rotations of an image) and construct our perturbation distribution to encapsulate these.
Generally speaking, the most important consideration is to ensure that the perturbation distribution places mass over \emph{\textbf{any}} test-time perturbations of concern; our results will later
demonstrate that we generally achieve good test-time robustness when we train with higher--variance perturbations than those we consider at test--time.

\section{THEORETICAL GENERALIZATION ANALYSIS} \label{genz}
The poor generalization properties of adversarial risk in high-dimensions are a fundamental limitation on its applicability and utility: regardless of the tractability of the optimization procedure of training, we are left with no guarantees (or even an expectation) that our classifier will be robust at test-time. We will now show that SRRs do not suffer from the same limitation. 

Given a neural network function class $\mathcal{F}$ and loss function class $L_\mathcal{F} \triangleq \{(X, Y) \to L(X, Y, f): f \in \mathcal{F}\}$, we can bound the generalization error of a classifier using the following theorem \citep{MohriEtAl2012}:
\begin{theorem} \label{yin}
Suppose $0 \leq L(X, Y, f) \leq c$ for all $X, Y, f$. Suppose further that the samples $S = \{(x_1, y_1), ..., (x_N, y_N)\}$ are i.i.d. from a distribution $p_\mathcal{D}$. Then for any $\delta \in (0, 1)$, with probability at least $1-\delta$ the following holds for all $f \in \mathcal{F}$:
\begin{align}\label{eq:theorem1}
\begin{split}
\vphantom{\sqrt{\log(2/\delta)}}r_\mathcal{D}(f)& - R_N(f) \\
&\leq 2c\;\textup{Rad}_S(L_\mathcal{F}) + 3c\sqrt{\log(2/\delta)/(2N)}.
\end{split}
\end{align}
\end{theorem}

This bound is probabilistic, data-dependent and uniform over all $f \in \mathcal{F}$. This means it holds for all $f \in \mathcal{F}$, including those trained on the dataset $S$. Informally, this means that with high probability (in the formal sense) the empirical risk on the training dataset will be ``close'' to the true risk (i.e. difference bounded by the term on the RHS).

To take advantage of this bound, we need to be able to compute $\textup{Rad}_S(L_\mathcal{F})$. 
The ERC (see Eq.~\ref{eq:emp_rad}) of the neural network function class $\textup{Rad}_S(\mathcal{F})$ can be upper bounded \citep{BartlettEtAl2017, YinEtAl2019} by an expression $O(\log(d_{\text{max}}))$ in dimension, where $d_{\text{max}}$ is the maximal number of nodes in a single layer. 
 Thus we simply need to relate $\textup{Rad}_S(L_\mathcal{F})$ to $\textup{Rad}_S(\mathcal{F})$.

Consider first the natural case, for which $L^{nat}(X, Y, f) \triangleq \phi(f(X), Y)$. If $\phi(\cdot, \cdot)$ is $\gamma-$Lipschitz in the first argument, we can use the Talagrand Contraction Lemma \citep{LedouxTalagrand1991}, which gives that $\textup{Rad}_S(L_\mathcal{F}) \leq \gamma \,\textup{Rad}_S(\mathcal{F})$. Thus, substituting this inequality into \eqref{eq:theorem1}, we have
\begin{align}\label{eq:natural_bound}
\begin{split}
\vphantom{\sqrt{\log(2/\delta)}}r_\mathcal{D}(f)& - R_N(f) \\
&\leq 2c\gamma \; \textup{Rad}_S(\mathcal{F}) + 3c\sqrt{\log(2/\delta)/(2N)}
\end{split}
\end{align}
such that our generalization error bound scales as $O(\log(d_{\text{max}}))$ in dimension (as $\textup{Rad}_S(\mathcal{F})$ is $O(\log(d_{\text{max}}))$).

We now introduce an analogous result for SRRs. In this case, the empirical risk we will use is the MC estimate $R_{N, C}^{\text{statMC}}$ in~\eqref{eq:MC} since this is what we actually compute. 

\begin{theorem}
Suppose $\phi$ is bounded in $[0, c]$, and $\gamma$-Lipschitz in the first argument. For $m\in\{1, ..., C\}$, define $S_m' = \{(x_{1, m}', y_1), ..., (x_{N, m}', y_N)\}$, such that it contains the $m$-th perturbed point from each of the $N$ original input points. For any $\delta\in(0,1)$, with probability at least $1-\delta$, the following holds for all $f\in\mathcal{F}$:
\begin{align}\label{eq:srr_bound2}
\begin{split}
\vphantom{\sqrt{\log(2/\delta)}}r_\mathcal{D}^{\text{stat}}(f)& - R_{N, C}^{\text{statMC}}(f) \\
&\leq 2c\gamma \,\overline{\textup{Rad}_{S'}(\mathcal{F})}
+ 3c\sqrt{\log(2/\delta)/(2N)}
\end{split}
\end{align}
where
\begin{align}
\overline{\textup{Rad}_{S'}(\mathcal{F})} \, \triangleq \frac{1}{C} \sum_{m = 1}^{C} \textup{Rad}_{{S'}_m}(\mathcal{F})
\end{align}
\end{theorem}
\begin{proof}
	See Appendix \ref{apx:proof}.
\end{proof}

Thus the SRR generalization error is upper bounded by an expression that varies as $O(\log(d_{\text{max}}))$. 

In contrast, for the adversarial risk, where (in binary classification) $L^{adv}(X, Y, f) \triangleq \max_{\delta \in \Delta}\phi(f_\theta(X+\delta), Y)$, $\textup{Rad}_S(L^{adv}_\mathcal{F})$ is \textit{\textbf{lower bounded}} by an expression containing explicit dependence on $\sqrt{d_{in}}$, where $d_{in}$ is the dimension of the input layer to the NN \citep{YinEtAl2019}. While this lower bound does not allow us to directly bound the generalization error using~\eqref{eq:theorem1}, it does suggest that in high dimensions the adversarial generalization error can be much greater than the natural and statistically robust generalization error. 
This indicates it will typically be difficult to train networks that are adversarially robust at test time for high-dimensional datasets. 
Our analysis thus shows that statistically robust networks may be easier to obtain.

\section{RELATED WORK}\label{sec:related-work}

\paragraph{Probabilistic robustness} Compared to the vast body of work on adversarial metrics for neural network robustness, there has been relatively little work examining robustness to probabilistic perturbations. \citet{FawziEtAl2018} introduced a risk metric based on finding the largest possible uniform perturbation distribution that still maintains a target level of accuracy.
\citet{HendrycksEtAl2019} experimentally evaluated different network architectures by averaging their accuracy over a discrete set of common image corruptions.
\citet{pmlr-v97-weng19a} and \citet{WebbEtAl2019}, suggested probabilistic metrics for \textit{\textbf{pointwise}} robustness based on verification and statistical sampling approaches respectively.
Our work extends these ideas to provide a comprehensive robust risk framework that applies to the whole network and which can be used for \emph{\textbf{training}} networks.

\paragraph{Use of noise in achieving adversarial robustness} 
Some recent papers~\citep{ZantedeschiEtAl2017,LiEtAl2019,CohenEtAl2019} have examined the use of noise/random corruptions as a mechanism for achieving adversarial robustness. For instance, randomized smoothing \citep{GilmerEtAl2019} can be used to harden modes against adversarial attack post-hoc with guarantees (see Appendix \ref{apx:randsmooth} for further comparison). Our work instead focuses on statistical robustness as the goal in its own right.

\paragraph{Distributional shift} Defining metrics for---and obtaining classifiers robust to---distributional shift from train to test is a related problem~\citep{CandelaEtAl2009, Duchi+Namkoong2018, LiptonEtAl2018}.
We instead are not assuming uncertainty in the population distribution, but that individual datapoints are probabilistically corrupted. 

\paragraph{Data augmentation for generalization} Training neural networks with randomly perturbed inputs is, of course, not a new concept. 
Many works examine this form of data augmentation as a means for improving generalization~\citep{elman1988learning,an1996effects}.
Other work has investigated training neural networks by perturbing other components of the neural network such as weights~\citep{an1996effects, graves2013speech}, targets~\citep{szegedy2016rethinking, vaswani2017attention}, and gradients~\citep{neelakantan2015adding}, with similar motivations.
\citet{ChapelleEtAl2001} introduced an empirical metric---vicinal risk---as a means to better approximate the true natural risk by using a kernelized density estimate for the data distribution $p_{\mathcal{D}}$, rather than just taking the standard MC approximation (empirical risk).
This leads to training schemes equivalent to randomly perturbing the inputs.

Our work differs from these in that training with random perturbations emerges from a principled risk minimization framework, rather than being taken as the starting point of algorithmic development.
Moreover, we use input perturbations not only during training but also as a means of evaluating the robustness at \textbf{\emph{test--time}}.
We have also drawn novel connections and comparisons between existing adversarial/robustness methods and probabilistic input perturbations, providing conceptual, theoretical, and empirical arguments for why the latter is an important component in the greater arsenal of robust classification approaches.

\section{EXPERIMENTS} 
\label{sec:experiments}

To empirically investigate our SRR framework, we now present  experiments comparing it with natural and adversarial approaches.
For SRR training, we follow the approach from Section \ref{esttra}, generating perturbations $X'_n$ to points $X_n$ in the training dataset and then using a mini-batch version of the gradient update in \eqref{grad}.
Unless otherwise stated, we train using the cross-entropy loss,
$
\phi^{\text{CE}}(f_\theta(X'), Y) \triangleq -\log(f_\theta(X')_Y),
$
referring to training on the resulting SRR as {\color{cyan}corruption training}.
Analogously to testing on accuracy in natural settings, we evaluate using the {\color{magenta}TSRM }, i.e.~\eqref{eq:total}.

\begin{figure}[t]
	\centering
	\includegraphics[width=\linewidth]{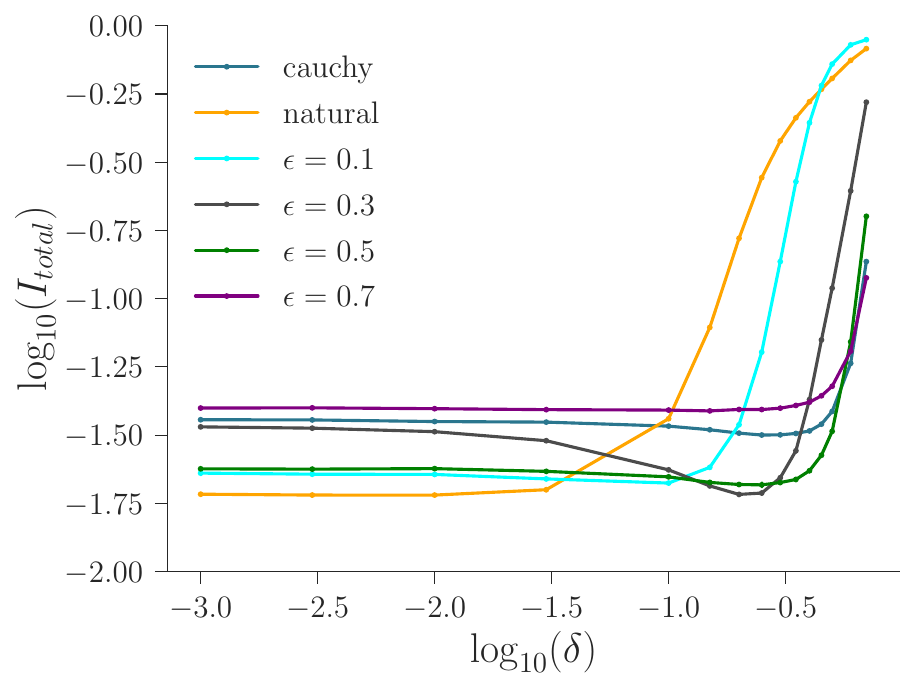}
	\caption{TSRM computed on the MNIST test set for different uniform perturbations $\epsilon$.
		Each line represents a different training objective.
		Results are averaged over 5 runs (error bars are imperceptibly small).}
	\label{fig:ccl}
\end{figure}

For Experiments \ref{natstat} and \ref{sec:tailor}, we used a dense ReLU network architecture with one hidden layer, while for Experiment \ref{sec:emp-gen}, we use a wide residual network architecture~\citep{ZagoruykoKomodaks2016}. Full details are provided in Appendix \ref{apx:imp}.

\begin{table*}[t!]
	\centering
	\caption{Train/test set evaluations of different networks on CIFAR-10, scores given in \% and averaged over 5 runs. The best test set performance for each evaluation metric is highlighted in bold.
	}
	{\normalsize
		\bgroup
		\def\arraystretch{1.1}
		\begin{tabular}{|l|l|p{1.5cm}|p{1.75cm}|p{1.75cm}|p{1.5cm}|}
			\hline
			\multicolumn{2}{|l|}{\multirow{2}{*}{}}                                                                        & \multicolumn{4}{c|}{\textbf{Training Method}} \\ \cline{3-6} 
			\multicolumn{2}{|l|}{}                                                                                         & \textcolor{orange}{Natural}, $\epsilon=0$ & \textcolor{cyan}{Corruption} $\epsilon=0.157$  & \textcolor{cyan}{Corruption} $\epsilon=0.5$ & \textcolor{brown}{PGD} ${\epsilon=0.157}$\\ \hline
			\multirow{4}{*}{\textbf{\begin{tabular}[c]{@{}l@{}}Evaluation\\ Metric\end{tabular}}} & {\color{olive} Natural}, $\delta=0$               &     98.7/\textbf{92.9}  &  $98.0$/$92.2$    &  $93.7$/$87.9$   &  $96.3$/$88.1$  \\ \cline{2-6} 
			& {\color{magenta} TSRM}, $\delta=0.157$ &    $94.9$/$89.4$      &                   98.1/\textbf{92.4}&      $94.1$/$88.1$     & $96.3$/$88.1$  \\ \cline{2-6} 
			& {\color{magenta} TSRM}, $ \delta=0.5$  &  $60.0$/$57.6$       &                   $79.9$/$76.0$& $95.6$/\textbf{89.9}   &  $94.9$/$86.1$     \\			\cline{2-6} 
			& {\color{red} Adversarial}, $\delta=0.157$               &  
			$0.0$/$0.0$ &  $0.2$/$0.2$    &  $3.1$/$3.0$ & $67.3$/\textbf{40.1}    \\ \hline
		\end{tabular}
		\egroup
	}
	\label{tabi}
\end{table*}

\subsection{Comparison to Natural Accuracy}\label{natstat}
\definecolor{col1}{rgb}{0.166617, 0.463708, 0.558119}
\definecolor{col4}{rgb}{0.3, 0.3, 0.3}
\definecolor{col5}{RGB}{0,128,0}
\definecolor{col6}{RGB}{128,0,128}
First, we show that naturally trained networks are vulnerable under the TSRM metric and that corruption training can alleviate this.
We train separate networks using 6 methods: corruption training with a \textcolor{col1}{Cauchy distribution} with scale $\gamma = 0.5$, corruption training with the uniform perturbations over radius-$\epsilon$ $L_\infty$ balls ($\epsilon = \textcolor{cyan}{0.1}, \textcolor{col4}{0.3}, \textcolor{col5}{0.5}, \textcolor{col6}{0.7}$), and \textcolor{orange}{natural training} ($\epsilon=0$).
We evaluate these networks using natural accuracy, and TSRM with 
 uniform perturbation distributions on radius-$\delta$ $L_\infty$ balls with $\delta$ from $10^{-3}$ to $0.7$.

\newcolumntype{P}[1]{>{\centering\arraybackslash}p{#1}}

The results, shown in Figure~\ref{fig:ccl}, provide several interesting insights. Firstly, as expected, networks corruption trained with more severe perturbations (larger $\epsilon$) performed better when evaluated on more severe perturbations (larger $\delta$), though this comes at the cost of a lower natural accuracy. Secondly, these gains are often more than an order of magnitude in size, confirming that TSRM can be highly distinct from natural accuracy ($\delta = 0$), and corruption training can provide significant benefits under this robust metric. Finally, training with a qualitatively distinct distribution (Cauchy) provided decent performance when evaluated on TSRM with uniform perturbations, supporting our intuition in Section~\ref{cpd}.

\subsection{Empirical Generalization Error}
\label{sec:emp-gen}

As previously noted, it has proved challenging to train networks to achieve high test-time adversarial accuracy on higher-dimensional datasets such as CIFAR-10 due to poor generalization from training.
By contrast, our analysis in Section \ref{genz} suggests that the gap will be more similar to natural accuracy for SRR approaches.
We thus investigate the generalization gap experimentally for TSRM on CIFAR-10.
Additionally, we compare corruption training with the PGD adversarial training method of \citet{MadryEtAl2018}, which is designed to maximize adversarial accuracy.

We train using four different methods: \textcolor{orange}{natural training}, \textcolor{cyan}{corruption training} with $\epsilon=0.157$ and  $\epsilon=0.5$, and \textcolor{brown}{PGD adversarial training} (7 gradient steps) with $\epsilon=0.157$.
Correspondingly, we then evaluate these networks using \textcolor{olive}{natural accuracy}, \textcolor{magenta}{TSRM} with $\delta = 0.157$ and $\delta = 0.5$, and \textcolor{red}{adversarial accuracy} with $\delta = 0.157$ (computed using 7-step PGD).
Here $0.157$ corresponds roughly 8/255 in pixel values, which is used as the corruption set by~\citet{MadryEtAl2018} for adversarial training, while $0.5$ represents a more extreme corruption model.

The results in Table~\ref{tabi} demonstrate generalization performance in line with our theoretical analysis: the \textcolor{olive}{natural}/\textcolor{magenta}{TSRM} evaluation metrics have fairly small generalization gaps (up to about $8\%$), while we see a much larger $27.2\%$ gap for \textcolor{red}{adversarial accuracy} (on the PGD trained network). To reiterate, this is a limitation of \textcolor{red}{adversarial risk} compared to \textcolor{magenta}{SRR}, \textit{\textbf{regardless of the training method}}.

Regarding training methods, for the \textcolor{olive}{natural} and \textcolor{magenta}{TSRM} metrics, we notice that the best test-set performance was achieved using the corresponding \textcolor{orange}{natural}/\textcolor{cyan}{corruption} training method. As can be expected, \textcolor{cyan}{corruption training} does not perform well on adversarial risk, since it targets SRR rather than adversarial risk. However, as discussed previously, adversarial risk is not a relevant or accurate metric to use for probabilistic perturbations. We also see that \textcolor{brown}{PGD training} is fairly effective for improving the TSRM, recording consistently good test-set TSRM for all values of $\delta$. However, there is still a consistent gap of $3-4\%$ between \textcolor{brown}{PGD training} and \textcolor{cyan}{corruption training} on these statistical risk metrics. This shows that adversarial risk can indeed be non-optimal for average-case performance, as we argued in Section \ref{sec:shortfall} (even if it can be a reasonable approximation). Further, \textcolor{brown}{PGD training} incurs significant additional computational expense ($\sim$6 times slower). We thus see that \textcolor{cyan}{corruption training} can be the better choice due to its simplicity and efficiency, when we are concerned with probabilistic perturbations.

\subsection{Tailored Loss Functions} \label{sec:tailor}
\begin{figure*}[!ht]
	\centering
	\settoheight{\tempdima}{\includegraphics[width=.363\linewidth]{figures/mnist_weighted_nat_sr}}%
	\begin{tabular}{@{}c@{ }c@{ }c@{ }}
		&\textbf{ \small \color{orange} Natural training} &\textbf{\small \color{cyan} SRR training} \\
		\rowname{\small \color{olive} Natural evaluation}& ~\includegraphics[width=0.363\linewidth]{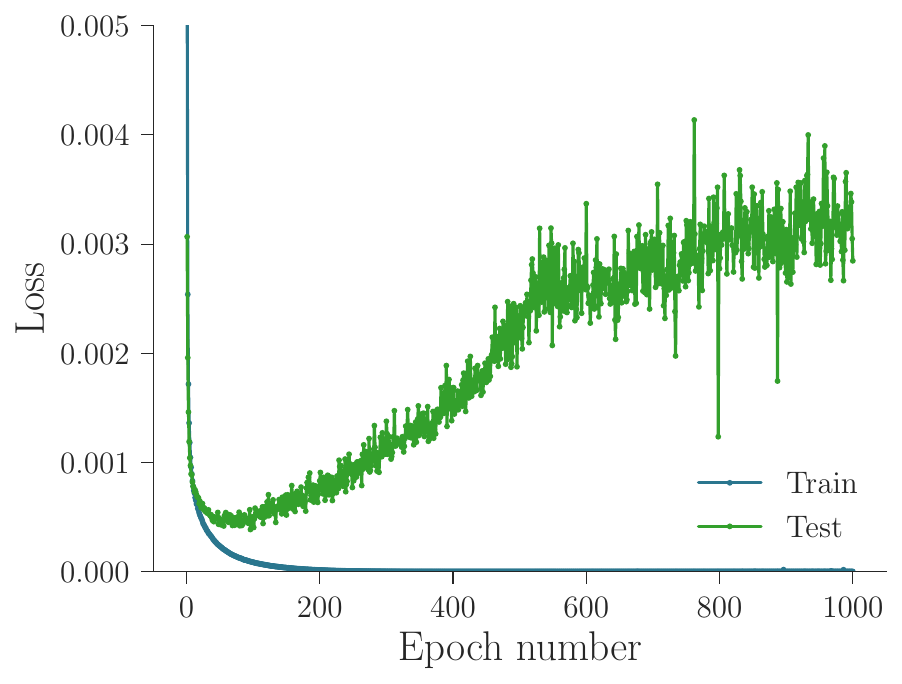}~~~~~~~~~~~~ & \includegraphics[width=0.363\linewidth]{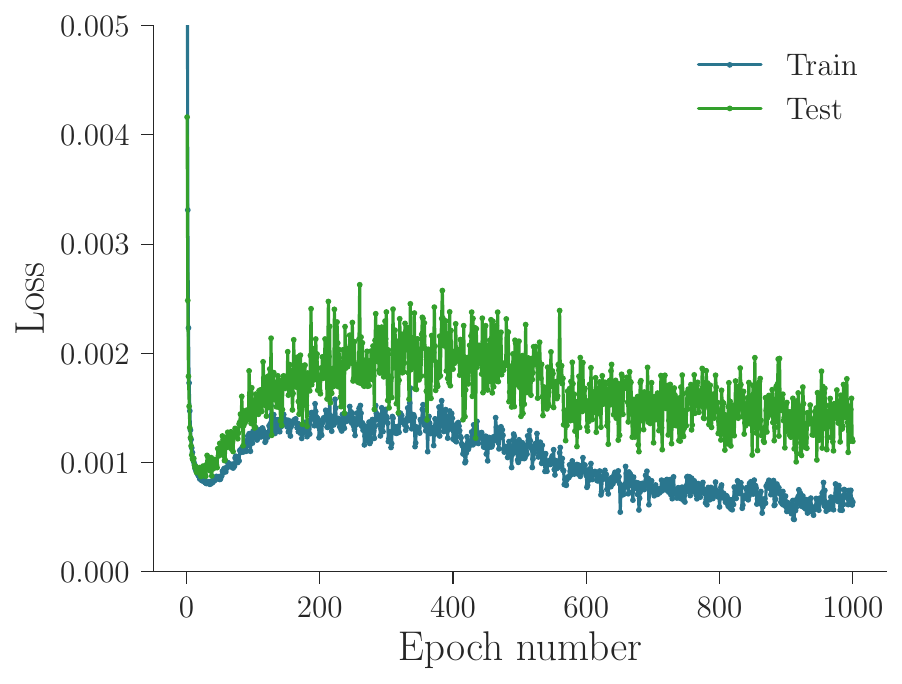} \\
		\rowname{\small \color{magenta} SRR evaluation}& ~\includegraphics[width=0.363\linewidth]{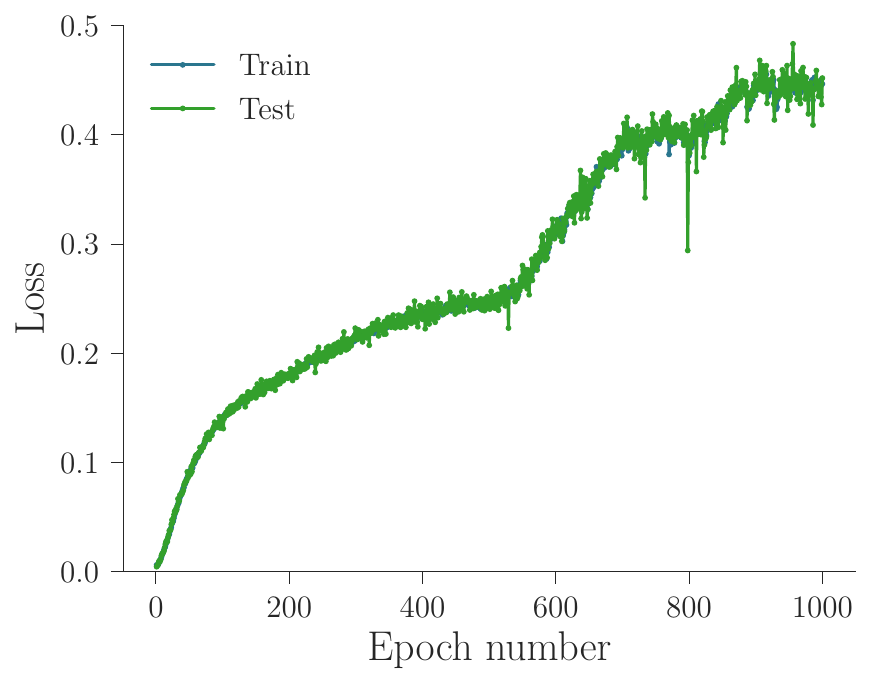}~~~~~~~~~~~~ & \includegraphics[width=0.363\linewidth]{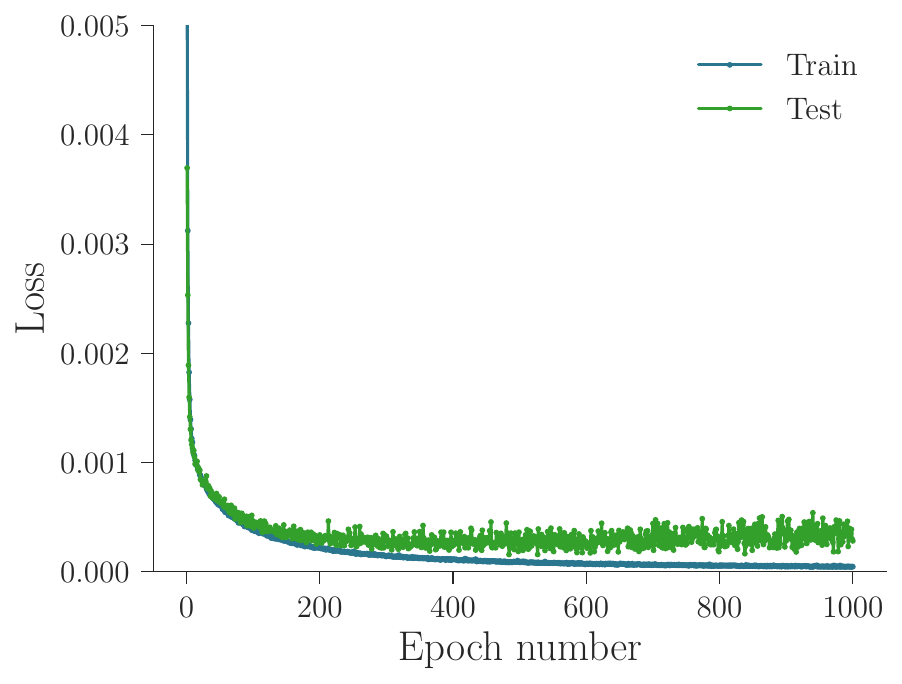} 
	\end{tabular}
	\caption{Learning curves using weighted cross-entropy loss on MNIST with $w(8) = 100$. Each plot represents a combination of training method and evaluation metric.
		For the bottom left plot, note the different y-axis scaling, and that the training and test SRR almost exactly overlap.
	}
	\label{fig:weighted}
\end{figure*}
In risk frameworks, we often wish to tailor the loss function $\phi$ to better represent a particular problem, such as a safety property.
For example, a self-driving car predicting the road is clear when there is actually a pedestrian will be far more damaging than predicting there is a pedestrian when the road is clear.
The SRR can be particularly useful in such situations, as networks need to be robust to noise in their inputs to fully incorporate all uncertainty present in the decision making process.
To demonstrate this, we consider training and evaluating using a weighted cross-entropy loss
\begin{align}
\phi(f_\theta(X'), Y) = - w(Y)\log f_\theta(X')_Y.
\end{align}
By taking $w(Y^*) >> 1$ for a particular problem class $Y^*$ and $w(Y) = 1$ for others, the classifier will be heavily penalized if it fails to correctly identify with high-confidence all occurrences of $Y^*$.
In turn, this heavy penalty can increase the sensitivity to perturbations in the inputs: the classifier should not confidently predict that $Y\neq Y^*$ if our input is close to points for which $Y=Y^*$, as this risks incurring the penalty if our inputs are noisy or our classifier is imperfect.

To make comparisons, we trained on MNIST using the same network as in Experiment \ref{natstat}, but with this weighted CE loss where $w(8) = 100$ and $w(Y)=1$ otherwise, i.e. penalizing classifiers which fail to confidently identify images of the number $8$. 
We also use a Gaussian perturbation distribution, taking $p(X'|X) \sim \mathcal{N}(X, \sigma^2 I)$ with $\sigma = 0.3$.  

The results, shown in Figure~\ref{fig:weighted}, exhibit several interesting traits.
Firstly, we see that {\color{orange} natural training} is extremely vulnerable to noisy input perturbations (bottom left), producing {\color{magenta} SRR values} at both train and test time that are multiple orders of magnitude worse than those achieved when {\color{cyan} corruption training} (bottom right).
This highlights both the importance of considering noisy inputs at test-time, and also the ability of SRRs to provide effective robust training.

Secondly, we see that while {\color{orange} training with natural risk} quickly overfits (top left), {\color{cyan} corruption training} with the SRR provides far better generalization (bottom right).
In fact, the final {\color{magenta} test SRR} of the corruption trained network is lower than the final {\color{olive} test natural risk} on the natural risk trained network, a powerful result given that the former evaluation metric is a corrupted version of the latter.
Thus even when the inputs are not corrupted, we can achieve a better loss by artificially adding noise to both the training procedure \textbf{and} at test-time.
This indicates that for the weighted loss, the SRR can provide robustness not only by accounting for potential input noise, but also by better accounting for the imperfect nature of the network to avoid overconfidently dismissing the potential for a test-time datapoint to belong to the problem class.

\section{CONCLUSIONS}
Motivated by applications where test-time corruptions are not generated adversarially but probabilistically, we introduced a statistically robust risk (SRR) framework, providing a class of metrics for evaluating robust performance under probabilistic input perturbations that are amenable to efficient training. 
We showed that SRRs can differ significantly from both natural and adversarial risk, and that networks with low test-time SRRs can be achieved through training with corrupted inputs.
Unlike adversarial risk, our results suggest that SRRs generalize from the training data similarly to, and potentially even better than, natural risks, meaning that they have more general practical applicability to high-dimensional datasets and complex architectures.
Thus, for probabilistic corruption threat settings, robust neural networks may be within reach for a wide range of applications.

\begin{acknowledgements}
TR gratefully acknowledges funding from Tencent AI
Labs and a Junior Research Fellowship supported by
Christ Church, Oxford.
\end{acknowledgements}

\bibliography{citations}

\clearpage
\appendix
\section{Proofs} \label{apx:proof}
\addtocounter{theorem}{-1}

Recall the risk definitions:
\begin{align} \tag{\ref{rstat}}
\hspace{-6pt}r^{\text{stat}}_\mathcal{D}(f_\theta) \triangleq \mathbb{E}_{(X, Y) \sim p_\mathcal{D}}\left[\mathbb{E}_{X' \sim p(\cdot|X)}\left[\phi(f_\theta(X'), Y) \right]\right]
\end{align}
\begin{align} \tag{\ref{eq:MC}}
	R^{\text{statMC}}_{N, C}(f_\theta) = \frac{1}{N} \sum_{n=1}^{N} \frac{1}{C} \sum_{m=1}^{C} \phi(f_\theta(x'_{n, m}), y_n) 
\end{align}

\begin{theorem}
Suppose $\phi$ is bounded in $[0, c]$, and $\gamma$-Lipschitz in the first argument. For $m = 1, ..., C$, define $S_m' = \{(x_{1, m}', y_1), ..., (x_{N, m}', y_N)\}$. In other words, $S_m'$ contains the $m$th perturbed point from each of the $N$ original input points. For any $\delta\in(0,1)$, with probability at least $1-\delta$, the following holds for all $f\in\mathcal{F}$:
\begin{align*}
\begin{split}
\vphantom{\sqrt{\log(2/\delta)}}r_\mathcal{D}^{\text{stat}}(f)& - R_{N, C}^{\text{statMC}}(f) \\
&\leq 2c\gamma \overline{\textup{Rad}_{S'}(\mathcal{F})}
+ 3c\sqrt{\log(2/\delta)/(2N)}
\end{split}
\end{align*}
where
\begin{align*}
\overline{\textup{Rad}_{S'}(\mathcal{F})} \triangleq \frac{1}{C} \sum_{m = 1}^{C} \textup{Rad}_{{S'}_m}(\mathcal{F})
\end{align*}
\end{theorem}
\begin{proof}
We can rewrite the SRR as a single expectation over $(X', Y)$ using the law of total expectation:
\begin{align*}
r^{\text{stat}}_\mathcal{D}(f) &= \mathbb{E}_{(X, Y) \sim p_\mathcal{D}, X' \sim p(\cdot|X)}\left[\phi(f_\theta(X'), Y) \right] \\
&= \mathbb{E}_{(X', Y) \sim q_\mathcal{D}}\left[\phi(f_\theta(X'), Y) \right]
\end{align*}
where $q_\mathcal{D}((X', Y)) \propto \int_{\mathcal{X}} p_\mathcal{D}(X, Y) p(X'|X) dX$.

For $C = 1$, we can apply the ERC generalization bound from Theorem \ref{yin} (on the distribution $q_\mathcal{D}$).

For $C > 1$ this is not directly possible because the $x'_{n, m}$ are not i.i.d. with respect to this distribution: for a fixed $n$, the $\{x'_{n, m}: m = 1, ..., C\}$ are dependent as they come from the same point $x_n$.

In the general case, we will use the idea of \textit{\textbf{independent blocks}} \citep{MohriEtAl2008}. That is, the fact that while the variables within each block $\{x'_{n, m}: m = 1, ..., C\}$ dependent, the blocks themselves are independent. To work at the block level, we need to rework our loss and risk definitions.

Recall the definition of the loss function class $L_\mathcal{F} = \{(X, Y) \to \phi(f(X), Y): f \in \mathcal{F}\}$.  Now we will define the "aggregate" loss function class $L_\mathcal{F}^C : \mathcal{X}^C \times \mathcal{Y} \to \mathbb{R}$ to include, for each function $l \in L_\mathcal{F}$, the following function:
\begin{align*}
	l'((x_{n, 1}, ..., x_{n, C}), y_n) = \frac{1}{C} \sum_{m = 1}^{C} l(x_{n, m}, y_n)
\end{align*}
That is, functions in this class compute the average loss for a neural network in $\mathcal{F}$ on the $C$ different points forming a block. 

Note that if $L_\mathcal{F}$ is bounded in $[0, c]$, then so is $L_\mathcal{F}^C$.
Further, notice that there is a 1-to-1 correspondence between NN functions $f \in \mathcal{F}$ and loss functions $l' \in L_\mathcal{F}^C$. Thus we can define the SRR and MC estimate of the SRR in terms of $l'$: $R^{\text{statMC}}_{N, C}(l') = \frac{1}{N} \sum_{n = 1}^N l'((x_{n, 1}, ..., x_{n, C}), y_n)$, and $r_\mathcal{D}^{\text{stat}}(l') = \mathbb{E}\left[R^{\text{statMC}}_{N, C}(l')\right]$, and these are the same as for the corresponding $f$.

As noted by \citet{MohriEtAl2008}, by viewing each block as an i.i.d. point and applying McDiarmid's inequality, the excess risk has the standard probabilistic bound. That is, with probability at least $1 - \delta$, for all $l' \in L_\mathcal{F}^C$:
\begin{align*}
\begin{split}
\vphantom{\sqrt{\log(2/\delta)}}r_\mathcal{D}^{\text{stat}}(l')& - R_{N, C}^{\text{statMC}}(l') \\
&\leq 2c \textup{Rad}_N(L_\mathcal{F}^C)
+ c\sqrt{\log(1/\delta)/(2N)}
\end{split}
\end{align*}

where $\textup{Rad}_N(L_\mathcal{F}^C)$ is the (non-empirical) Rademacher complexity of $L_\mathcal{F}^C$ over $N$ inputs. 

Once again applying McDiarmid's inequality in the standard way, we can convert to a bound involving the empirical Rademacher complexity over the sample $S' = \{((x_{1, 1}, ..., x_{1, C}), y_1), ..., ((x_{N, 1}, ..., x_{N, C}), y_N)\}$:
\begin{align*}
\begin{split}
\vphantom{\sqrt{\log(2/\delta)}}r_\mathcal{D}^{\text{stat}}(l')& - R_{N, C}^{\text{statMC}}(l') \\
&\leq 2c \textup{Rad}_{S'}(L_\mathcal{F}^C)
+ 3c\sqrt{\log(2/\delta)/(2N)}
\end{split}
\end{align*}

It remains to determine $\textup{Rad}_{S'}(L_\mathcal{F}^C)$. 

\begin{lemma}
$\textup{Rad}_{S'}(L_\mathcal{F}^C) \leq \frac{1}{C} \sum_{m = 1}^{C} \textup{Rad}_{{S'}_m}(L_\mathcal{F})$
\end{lemma}

\begin{proof}
Consider the following function class $U_\mathcal{F}^C: \mathcal{X}^C \times \mathcal{Y} \to \mathbb{R}$:
\begin{align*}
\begin{split}
U_\mathcal{F}^C((x_{n, 1}&, ..., x_{n, C}), y_n) \\
&= \{\frac{1}{C}\sum_{m = 1}^{C} l_m(x_{n, m}, y_n) : l_1, ..., l_C \in L_\mathcal{F}\}
\end{split}
\end{align*}

Since $U_\mathcal{F}^C$ is a linear combination of $C$ classes $L_\mathcal{F}$ divided by $C$, its ERC is given by $\textup{Rad}_{S'}(U_\mathcal{F}^C) = \frac{1}{C} \sum_{m = 1}^{C} \textup{Rad}_{S'_m}(L_\mathcal{F})$. This can be seen as follows:
\begin{align*}
\textup{Rad}_{S'}&(U_\mathcal{F}^C) = \frac{1}{N} \mathbb{E}_\sigma \left[ \sup_{u \in U_\mathcal{F}} \sum_{n = 1}^{N} \sigma_n u((x_{n, 1}, ..., x_{n, C}), y_n) \right] \\
&= \frac{1}{N} \mathbb{E}_\sigma \left[ \sup_{l_1, ..., l_C \in L_\mathcal{F}} \sum_{n = 1}^{N} \sigma_n \frac{1}{C} \sum_{m = 1}^{C} l_m(x_{n, m}, y_n) \right] \\
&= \frac{1}{N} \mathbb{E}_\sigma \left[\sup_{l_m \in L_\mathcal{F}} \frac{1}{C} \sum_{m = 1}^{C} \left( \sum_{n = 1}^{N} \sigma_n l_m(x_{n, m}, y_n) \right) \right] \\
&= \frac{1}{C} \sum_{m = 1}^{C} \frac{1}{N} \mathbb{E}_\sigma \left[\sup_{l \in L_\mathcal{F}} \sum_{n = 1}^{N} \sigma_n l(x_{n, m}, y_n) \right] \\
&= \frac{1}{C} \sum_{m = 1}^{C} \textup{Rad}_{S'_m}(L_\mathcal{F})
\end{align*}

Further, since $L_\mathcal{F}^C$ is a subset of $U_\mathcal{F}^C $ (the former is $U_\mathcal{F}^C$ with the restriction that all the constituent loss functions are the same), it has ERC at most $\textup{Rad}_{S'}(U_\mathcal{F}^C)$.
\end{proof}

 The result follows using the Ledoux-Talagrand contraction lemma to bound the ERC of the loss function class in terms of the ERC of the neural network function class.

\end{proof}

\section{Effect of Statistically Robust Training} \label{apx:effect}

In Section \ref{sec:shortfall}, we discussed one of the major shortfalls of adversarial risk: namely, that it loses information about how robust points are. We now demonstrate that this can be seen in practice with trained networks.

\begin{figure*}[!ht]
    \centering
    \begin{minipage}{.5\textwidth}
        \centering
        \includegraphics[scale=0.55]{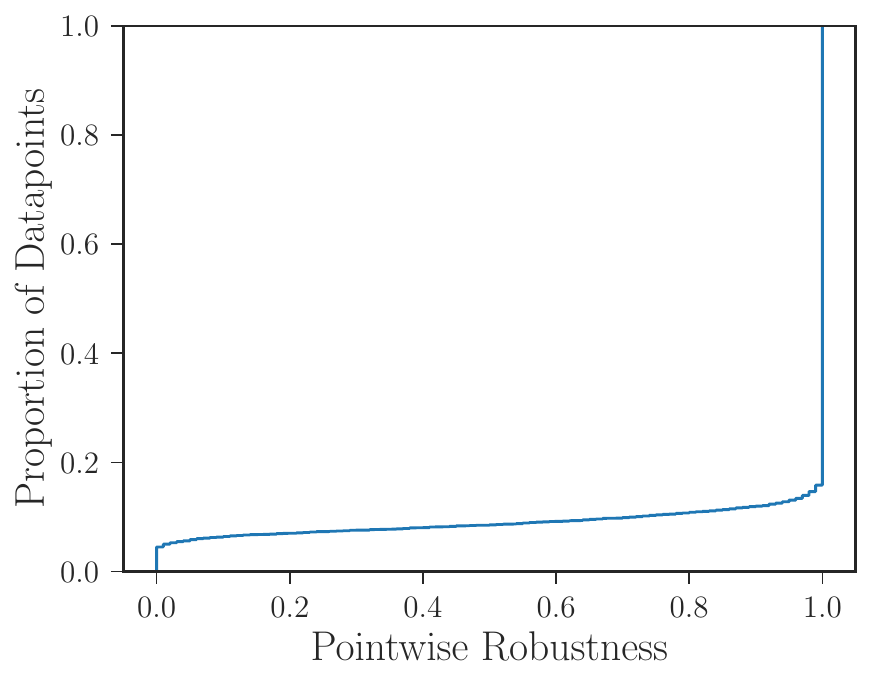}
        (a) Corruption trained ($\epsilon=0.157$)
    \end{minipage}%
    \begin{minipage}{0.5\textwidth}
        \centering
        \includegraphics[scale=0.55]{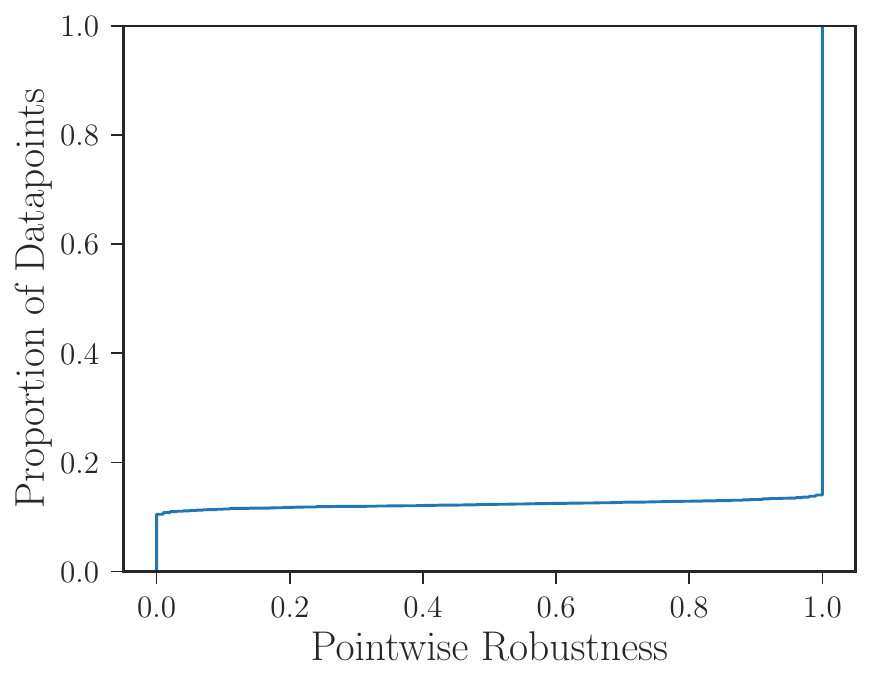}
        (b) Adversarially trained ($\epsilon=0.157$)
    \end{minipage}
    
    \caption{Empirical cumulative distribution function (ECDF) for the pointwise statistical robustness (CI definition), over all 10000 images in the CIFAR-10 test dataset.}
    \label{fig:points_cdf}
\end{figure*}

We analyse networks trained on CIFAR-10 either using corruption training (uniform over $\epsilon=0.157$ $L_\infty$ ball) or PGD/adversarial training (over $\epsilon$ $L_\infty$ ball). As we saw in Section \ref{sec:emp-gen} and Table \ref{tabi}, the former attains $92.4\%$ on the TSRM metric, while the latter attains $88.1\%$. We thus wish to examine to what the additional robust performance of the corruption trained network is attributable to. 

Recall that the pointwise statistical robustness of a point $(x, y)$ can be defined as $PSR(x, y) = \mathbb{E}_{X' \sim p(\cdot|x)}\left[\mathbbm{1}_{f_\theta(x') \neq y}\right]$ (note that, as throughout this paper, we have switched to using the CI rather than PC definition of adversarial examples). This expresses how robust the point is; if this is $0$, the network almost always predicts incorrectly, while if it is $1$, it almost always predicts correctly. In Figure \ref{fig:points_cdf}, we plot the empirical CDF of $PSR(x, y)$ over the 10000 points in the CIFAR-10 dataset, for both trained networks. 

We see that both networks have $PSR(x, y) = 1$ for the majority of points (more than $80\%$). The main difference, then, is in that fact that the corruption trained network has a large number of points with $PSR(x, y)$ between $0$ and $1$. In contrast, the adversarially trained network is very polarized, with the vast majority of points having $PSR(x, y) = 0$ or $1$. That is to say, the former achieves better TSRM not by being fully robust on more points, but rather by ensuring as many points as possible have \textit{some} robustness. Returning to the medical imaging analogy in Section \ref{sec:shortfall}, this means correct diagnoses for more patients, under random noise corruptions.

\section{Estimation and Training} \label{apx:esttra}
In Section \ref{esttra}, we discussed a simple Monte Carlo estimation/training scheme, which involves sampling $C$ perturbed points around each input:
\begin{align} \tag{\ref{eq:MC}}
	R^{\text{statMC}}_{N, C}(f_\theta) = \frac{1}{N} \sum_{n=1}^{N} \frac{1}{C} \sum_{m=1}^{C} \phi(f_\theta(x'_{n, m}), y_n) 
\end{align}

\subsection{Effect of $C$}
As we justified in Section \ref{esttra}, taking $C=1$ is, somewhat surprisingly, sufficient to estimate the SRR accurately, meaning that we can evaluate test loss/accuracy for a network by simply perturbing each sample once. We found that this also holds true for training: rather than taking multiple samples around each datapoint, as is often done in e.g. Gaussian data augmentation, we find that perturbing each sample once doing training is sufficient, achieving similar performance to taking more samples (e.g. $C = 5$) while reducing training time to the same as standard training. 

\subsection{Alternative Methods}

Nonetheless, other schemes may sometimes be more efficient in different situations, such as when our dataset is small, or when the number of "non-robust" points is small.

For example, suppose that we wish to evaluate the TSRM ($\phi(f_{\theta}(X'),Y)=\mathbbm{1}_{f_\theta(X') \neq Y}$) given a relatively small number of datapoints, say $N=100$. This can be achieved using our previous Monte Carlo method:
\begin{align} \tag{\ref{eq:MC}}
	R^{\text{statMC}}_{100, C}(f_\theta) = \frac{1}{100} \sum_{n=1}^{N} \frac{1}{C} \sum_{m=1}^{C} \phi(f_\theta(x'_{n, m}), y_n) 
\end{align}

However, since $N$ is much smaller, the law of large numbers over $N$ (datapoints) no longer applies and we must estimate the robustness of each point precisely:
\begin{align} \tag{\ref{eq:lstat}}
L^{\text{stat}}(X, Y, f) &\triangleq \mathbb{E}_{p(X'|X)}\left[\phi(f_\theta(X'), Y) \right]
\end{align}

For TSRM, $0 \leq L^{\text{stat}}(x_n, y_n, f) \leq 1$ for all $n$. However, many such points will have $L^{\text{stat}}(x_n, y_n, f)$ close to 0 or 1 (i.e. almost all of the perturbation region is classified correctly, or almost all is classified incorrectly), in which case sampling from $p(\cdot|x_n)$ to estimate this quantity will have very low variance. Thus it can be more efficient not to uniformly increase the number of samples for each point $x_n$ by increasing $C$ in the Monte Carlo estimator, but to adaptively select which points to sample from based on the variance observed thus far.

\begin{figure}[t]
	\centering
	\includegraphics[width=0.9\linewidth]{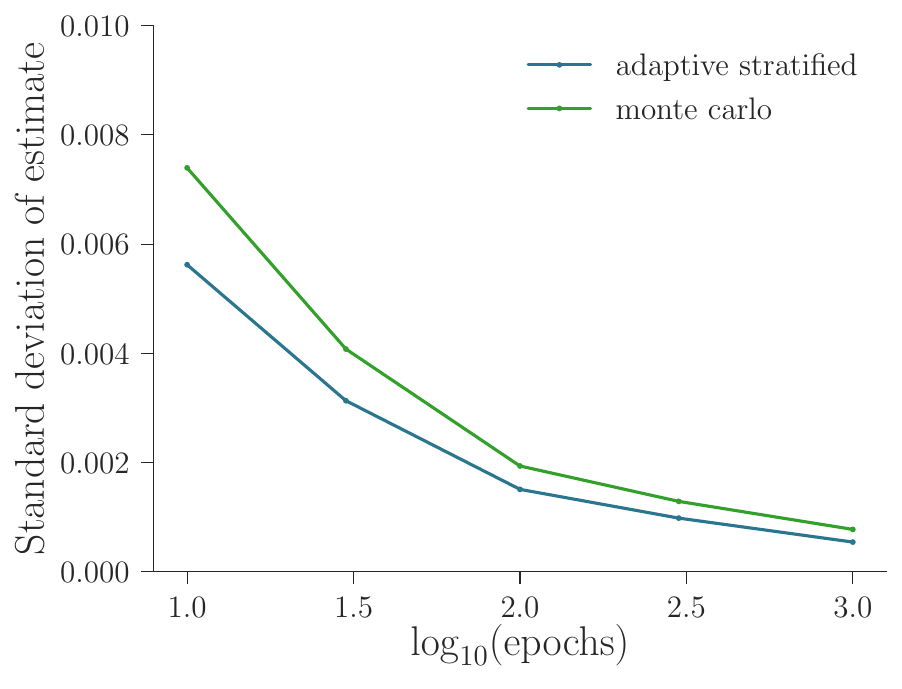}
	\caption{Standard deviation over 100 runs of the adaptive stratified and Monte Carlo algorithms for estimating TSRM. The horizontal axis represents the number of forward passes allowed.}
	\label{fig:est}
\end{figure}

This is a stratified sampling problem which has connections to multi-armed bandits, where the ultimate goal is to produce a low-variance estimate of $r^{\text{stat}}(p,f_\theta)$. \citet{CarpentierEtAl2015} suggested an algorithm (\textit{\textbf{adaptive stratified sampling}}) for this purpose. In Fig \ref{fig:est} we investigated the variance of adaptive
stratified sampling for estimating TSRM, and found that
it was somewhat lower given the same number of epochs
(i.e. same number of forward passes using the network)
compared to Monte Carlo. For larger datasets, however,
adaptive sampling provides no noticeable advantage, which
is why we employ Monte Carlo in our other experiments.

Alternative methods for training networks with respect to
the SRR are also possible, and also could potentially be beneficial particularly when we have limited data. To this
end we applied an importance sampling scheme to the setting
in Experiment \ref{sec:tailor}, but found that this did not provide noticeable improvements. Investigation of adaptive training methods, perhaps borrowing ideas from active learning,
could comprise interesting avenues for future work.

\section{Comparison to Randomized Smoothing} \label{apx:randsmooth}

Randomized smoothing is a recently proposed technique that applies post-hoc to a classifier in order to improve its adversarial robustness. Given a base classifier $f$, a new "smoothed classifier" $g$ is defined as follows:
\begin{align} \label{eqn:rand_smooth}
    g(x) \triangleq \argmax_{c} \mathbb{P}_{X' \sim p(\cdot|x)}(f(X') = c)
\end{align}
Here, $p(\cdot|x)$ is some perturbation distribution, usually taken to be additive Gaussian noise, i.e. $X' \sim N(x, \sigma^2 I)$. For a given input point $x$, $g$ picks the class $c$ which is predicted most often by $f$ in the perturbation region. In practice, since it is not possible to evaluate predictions over the whole distribution, we sample from the distribution, and choose the class which is predicted most frequently (majority vote). This naturally leads to more smooth/invariant predictions, and in fact it is possible to directly obtain certified adversarial robustness (accuracy) guarantees in $l_2$ norm for each point: that is, a radius $R_x$ such that:
$$g(x_{\text{pert}}) = g(x) \;\;\;\;\;\;\;\;\; \forall x_{\text{pert}} \text{ s.t. } ||x_{\text{pert}} - x||_2 < R_x$$
The size of this radius $R_x$ depends on both the probabilities of the most likely classes predicted by $f$ in $N(x, \sigma^2 I)$, as well as the standard deviation $\sigma$ of the Gaussian distribution. Intuitively, if $x_{\text{pert}}$ is sufficiently close to $x$, then the distributions $N(x_{\text{pert}}, \sigma^2 I)$ and $N(x, \sigma^2 I)$ overlap sufficiently that the most likely predictions of $f$ on both are the same. There exists a tradeoff between smoothness and precision here: increasing $\sigma$ increases the certified adversarial radius, but can also make the classifier "too smooth" and lose standard accuracy. It can be seen that in the limit $\sigma \to \infty$, $g$ will simply predict the same class for all input points $x$. 

At first glance, the smoothing operation (\ref{eqn:rand_smooth}) appears to resemble the pointwise statistical robustness metric from Section \ref{statrob}:
\begin{align} \tag{\ref{eq:psr}}
\begin{split}
\mathcal{I}[p] &\triangleq 
\mathbb{E}_{X' \sim p(\cdot|x)} \left[\mathbbm{1}_{f(X') \neq f(x)}\right]
\end{split}
\end{align}
as both involve probabilistic perturbations around point $x$. However, notice that the former is an operation, which produces a new classifier, while the latter is a metric for the original classifier $f$. In particular, the smoothing $f \to g$ does not directly minimize or target the statistical robustness metric in any meaningful way. That said, we could of course apply the metric to $g$, which would involve a "double averaging". The inner component of (\ref{eq:psr}) would be $\mathbbm{1}_{g(X') \neq g(x)}$, which tests whether the smoothed classifier classifies $X', x$ the same, or, in other words, whether $f$ agrees sufficiently on $N(X', \sigma^2 I)$ and $N(x, \sigma^2 I)$. For similar reasons to the adversarial robustness guarantees above, we would expect the pointwise statistical robustness metric to be small on $g$ (in fact, $0$ if $p(\cdot|x)$ is chosen such that its support lies entirely within the $l_2$ ball of radius $R_x$ around $x$).

We now consider our total statistical robustness metric defined in Section \ref{sec:tsrm}, which extends to the whole data distribution $p_\mathcal{D}$. Recall that, crucially, when moving to the TSRM from the pointwise metric, we changed to the "corrupted instance" (CI) definition of adversarial examples; that is, we compare the perturbed prediction to the true class label, rather than the original prediction:
\begin{align}\tag{\ref{eq:total}}
\mathcal{I}_{\text{total}}[p]
&= \mathbb{E}_{(X, Y) \sim p_\mathcal{D}}\left[\mathbb{E}_{X' \sim p(\cdot|X)}\left[\mathbbm{1}_{f(X') \neq Y}\right]\right],
\end{align} 

The CI definition requires that the classifier is not just "smooth" in the sense of producing the same prediction at nearby points, but also that this prediction matches the true label $Y$. Now consider applying randomised smoothing to the base classifier $f$, in addition to or in lieu of SRR training. As before, the randomized smoothed classifier $g$ does not directly target TSRM. However, unlike the pointwise metric, we would not necessarily even expect $g$ to obtain a better TSRM than $f$. This is because $g$ makes $f$ smoother, but this can come at the cost of making the correct prediction.

We empirically tested applying randomized smoothing to our classifiers, and as we expected, this provided no benefit in terms of SRR/TSRM, usually performing slightly worse on these metrics. The smoothed classifiers did exhibit greater adversarial robustness, as randomized smoothing is designed to achieve. Thus, while randomized smoothing makes use of probabilistic perturbations, it does so in a very different way to SRR training. 

\section{Implementation Details} \label{apx:imp}
All experiments were performed using Python 3 with the PyTorch framework, using a single NVIDIA Tesla T4 GPU.

In Experiment \ref{natstat}, we used the MNIST dataset with the standard train/test split (60000/10000), with pixels scaled to $[0, 1]$. We used a dense ReLU network architecture with an input layer of size 784, a hidden layer of size 256, and an output layer of size 10. Training was performed for 50 epochs, each time using default initialization and with the Adam optimizer with learning rate $1\mathrm{e}{-3}$.

In Experiment \ref{sec:emp-gen}, we used the CIFAR-10 dataset with the standard train/test split (50000/10000), with pixels normalized by per-channel mean and standard deviation. We use a wide residual network architecture~\citep{ZagoruykoKomodaks2016} with depth 28, widening factor 10, and dropout rate 0.3. Training was performed using SGD with a staggered learning rate, starting at $0.01$ and ending at $0.0004$. We report the train/test scores after 30 epochs (averaged over 5 runs) in Table~\ref{tabi}. Adversarial training was applied using 7-step PGD to find the most adversarial perturbation for each training point; this took significantly longer (approx. $6$ times) compared to natural or corruption training.

In Experiment \ref{sec:tailor}, we again used the MNIST dataset with the same train/test split and the same network architecture. However, we instead trained for 1000 epochs in each run, and used a learning rate of $5\mathrm{e}{-5}$ for the Adam optimizer.

\end{document}